\documentclass{article}

\PassOptionsToPackage{numbers, compress}{natbib}
\usepackage[preprint]{neurips_2020}

\usepackage[utf8]{inputenc} % allow utf-8 input
\usepackage[T1]{fontenc}    % use 8-bit T1 fonts
\usepackage{hyperref}       % hyperlinks
\usepackage{url}            % simple URL typesetting
\usepackage{booktabs}       % professional-quality tables
\usepackage{amsfonts}       % blackboard math symbols
\usepackage{nicefrac}       % compact symbols for 1/2, etc.
\usepackage{microtype}      % microtypography

\usepackage{graphicx}
\usepackage{comment}
\usepackage{amsmath}
\usepackage{amssymb} % define this before the line numbering.
\usepackage{color}
\usepackage{multirow}
\usepackage{multicol}
\usepackage{bm}

\newlength\savewidth\newcommand\shline{\noalign{\global\savewidth\arrayrulewidth
  \global\arrayrulewidth 1pt}\hline\noalign{\global\arrayrulewidth\savewidth}}

\title{PriorGAN: Real Data Prior for Generative Adversarial Nets} % Replace with your title

\author{
Shuyang Gu$^{1}$, Jianmin Bao$^{2}$\thanks{Corresponding author.}, Dong Chen$^{2}$,
Fang Wen$^{2}$ \\
$^1$\emph{University of Science and Technology of China}\\
$^2$\emph{Microsoft Research} \\
gsy777@mail.ustc.edu.cn\; \{jianbao, doch, fangwen\}@microsoft.com
}

\begin{document}

\maketitle
\vspace{-0.3cm}
\begin{abstract}
%Generative adversarial networks (GANs) have achieved impressive results today, but are still not good enough for two reasons: quality of images are not good enough and it may mode dropping from real data distribution. To improve these two issues, in this paper, we build a parametric prior of real data distribution, using this prior help us generate more releastic and diverse images.

%Generative adversarial networks (GANs) have achieved rapid progress in learning rich data distributions. The data distribution is usually learned by an adversarial training process of a generator and a discriminator. However, the discriminator is always trained with a small batch of the real data, which may mislead the decision boundary that causes poor performance of the generator. To this end, we propose a novel prior that captures the whole real data distribution. This prior have a global sense of the real data distribution that prevent the misleading of the discriminator. This prior can be flexibly applied in current GAN frameworks, which are called PriorGAN. To get the prior of data, we adopt a simple yet elegant Gaussian mixture model(GMM) to build an explicit probability distribution for the whole real data. Then we apply the model to restrict the generated samples to fit the prior. Our experiments demonstrate that our proposed PriorGAN ourperforms state-of-the-art GAN frameworks(PGGAN, StyleGAN) on FFHQ, LSUN cat, LSUN car, and other datasets by large margins.

%  Imprecise boundary problem and missing modes problem

Generative adversarial networks (GANs) have achieved rapid progress in learning rich data distributions. However, we argue about two main issues in existing techniques. First, the \emph{low quality} problem where the learned distribution has massive low quality samples. Second, the \emph{missing modes} problem where the learned distribution misses some certain regions of the real data distribution. To address these two issues, we propose a novel prior that captures the whole real data distribution for GANs, which are called PriorGANs. To be specific, we adopt a simple yet elegant Gaussian Mixture Model (GMM) to build an explicit probability distribution on the feature level for the whole real data. By maximizing the probability of generated data, we can push the low quality samples to high quality. Meanwhile, equipped with the prior, we can estimate the missing modes in the learned distribution and design a sampling strategy on the real data to solve the problem. The proposed real data prior can generalize to various training settings of GANs, such as LSGAN, WGAN-GP, SNGAN, and even the StyleGAN. Our experiments demonstrate that PriorGANs outperform the state-of-the-art on the CIFAR-10, FFHQ, LSUN-cat, and LSUN-bird datasets by large margins.

\end{abstract}

\vspace{-0.3cm}
\section{Introduction}
\vspace{-0.3cm}
% Recent studies have shown remarkable success in generative models for their wide applications like high quality image generation, image-to-image translation, and so on. However, there is still a gap between the best generation results and real data[]. We found that in current GAN models, there are two main issues which may cause this gap: imprecise boundary problem and modes missing (modes dropping) problem. Therefore, this work proposes a novel method PriorGAN to solve these problems.

%Recent studies have shown remarkable success in generative models for their wide applications like high quality image generation, image-to-image translation, and so on. We usually evaluate a GAN in two perspectives: \emph{Quality} and \emph{Diversity}. Specifically, on the one hand, we want GAN to generate high quality images, it means each generated image should look like real images, on the other hand, for each image in real data distribution, we need our GAN to have the capbility to generate a similar image.

Generative Adversarial Networks (GANs)~\cite{goodfellow2014generative} have achieved great success in learning high dimensional probability distributions. The capability of GANs advances many important and useful applications, like high quality image generation~\cite{karras2017progressive,brock2018large}, image-to-image translation~\cite{isola2017image,wang2018high}, image editing~\cite{perarnau2016invertible, gu2019mask}, and so on. However, GANs still suffer from the unstable training process.

Many recent works~\cite{radford2015unsupervised,mao2017least,arjovsky2017wasserstein,gulrajani2017improved,miyato2018spectral} focus on stabilizing the training process of GANs. Despite their success, there are still two main issues in current GANs. First, the learned distribution always contains massive low quality samples. We refer to this phenomenon as the \emph{low quality} problem. Second, the learned distribution misses certain areas of real data distribution. We refer to this phenomenon as the \emph{missing modes} (also known as mode dropping) problem. We have carefully studied these two problems and observed that they may be caused by the inaccurate gradient direction from the discriminator. 

Ideally, the gradient direction of the discriminator to the generator will make the generated data distribution as close to the real data distribution as possible. However, we found that in practice, the discriminator cannot achieve the theoretical optimal point due to insufficient iteration number or limited fitting ability. In this situation, the gradient direction given by the discriminator is sometimes inaccurate. Previous methods~\cite{gulrajani2017improved,miyato2018spectral} try to solve the inaccurate gradient problem with the Lipschitz constraint for the discriminator. But the Lipschitz constraint will further limit the capability of the discriminator so that it still can not provide accurate gradient for the generated data. More recent work LGAN~\cite{zhou2019lipschitz} proved that penalizing Lipschitz constant
guarantees the gradient for generated data is accurate, but it requires the discriminator need to be optimal before each iteration of generator training, which is impractical for GANs training.

%For better understanding, we show two typical situations on toy examples in Figure~\ref{fig:toy_vis}, The green points are the real data, the blue points are the generated data, the red arrow represents the gradient direction of the discriminator for generated data. In Figure~\ref{fig:toy_vis}(a), some generated data are far from the real image distribution. But the gradient for these low quality generated points (out of the real data distribution) is quite random, therefore they can't move to the real data distribution and stay in the situation of low quality problem. In Figure~\ref{fig:toy_vis}(c), we notice that the real data distribution has two main regions, but the gradient for the generated points have the same direction to one region of the real data, this seems to be an intrinsic cause of the missing modes problem\footnote{For more experiment detail, please refer to the supplemental material.}.

To address these issues, we adopt an auxiliary model to help the generated data get the accurate gradient direction. We propose to build an explicitly prior to the real data distribution for GANs, which we call PriorGANs. Specifically, we choose to adopt the Gaussian Mixture Model (GMM) to establish a probabilistic model for the real data distribution. Equipped with this model, we can address both issues. For the low quality issue, we can apply the model to estimate the probability of each generated image, the probability can also indicate image quality. By encouraging the low quality samples to get a higher probability in the model, low quality samples will get a gradient direction to the real data nearby. Therefore the low quality problem could be solved. We call this the \emph{quality loss}.  %as illustrated in Figure~\ref{fig:toy_vis}(b). 

For the missing modes issue, we can take full advantage of the GMM model to divide the real data into several groups. Each group may represent a specific mode. Then we can get the densities of the real data in each group. During the training process of GANs, we can estimate the densities of the generated data in each group. If the density of generated data in a group is lower (or higher) than real data, we will increase (or decrease) the sampling frequency of the real data in that group for the next training iteration. We call this the \emph{resampling strategy}. It will help the generative model solve the missing modes problem.% as illustrated in Figure~\ref{fig:toy_vis}(d).

The proposed PriorGAN is simple, flexible, and general. Our experiments demonstrate that it is suitable for various types of GANs, and it outperforms state-of-the-art GAN frameworks, including LSGAN, WGAN-GP, SNGAN and StyleGAN, on CIFAR-10, FFHQ, LSUN-cat and LSUN-bird datasets by large margins.

In summary, our key contributions are:

\begin{enumerate}
\vspace{-0.1cm}
    \item We demonstrate the \emph{low quality} and \emph{missing modes} problems in current GANs and demonstrate the cause of these two problems on toy examples.
    \vspace{-0.1cm}
    \item We propose \emph{real data prior} for GANs(PriorGANs), equipped with the prior, we propose the \emph{quality loss} and \emph{resampling strategy} help current GANs solve these problems.
    \vspace{-0.1cm}
    \item Our experiments demonstrate that PriorGANs outperform state-of-the-art GAN frameworks on various datasets by large margins.
    \vspace{-0.1cm}
\end{enumerate}

\vspace{-0.3cm}
\section{Related Work}
\vspace{-0.3cm}
Over the past years, GANs have achieved significant attention in the deep learning community. Massive works are proposed to improve the performance of GANs. In this section, we will briefly review them. Also, we will review some classical algorithms about modeling data distribution.

\noindent \textbf{Generative Adversarial Nets}.
Early works about GANs try to solve the unstable training process of GANs, many works focus on proposing new loss functions to solve this problem. For example, LSGAN~\cite{mao2017least} replace the original BCE loss with the proposed least square loss. WGAN~\cite{arjovsky2017wasserstein} proposes the Wasserstein distance and weight clipping to achieve the Lipschitz condition. Later work began to notice the Lipschitz condition. Some typical works are as follows:
WGAN-GP~\cite{arjovsky2017wasserstein} further introduce the gradient penalty to achieve the Lipschitz condition. SNGAN~\cite{miyato2018spectral} proposes Spectral Normalization to achieve Lipschitz condition for training GANs. More recent works~\cite{dong2019margingan,NIPS2019_8560,cao2019multi} design task-specific loss functions for GANs. Meanwhile, some works aim to improve the diversity of the GANs, for instance, VEEGAN~\cite{srivastava2017veegan} and CVAE-GAN~\cite{bao2017cvae} proposes to reconstruct the images in the pixel level to avoid the missing modes problems. MS-GAN~\cite{mao2019mode} proposes a loss term that maximizes the ratio of the distance between generated images with respect to the corresponding latent codes. Besides new loss functions, there are also some works~\cite{radford2015unsupervised, karras2017progressive, zhang2018self, karras2019style} focus on using different network structures and training strategies for training. In contrast to these methods, our method proposes to build explicit real data prior to solving these problems.

%\textbf{Improve GANs from architectures perspective.} 

%\textbf{Improve GANs from other perspectives.} 

%Recent studies have shown remarkable success in GANs. These works can be separated into three perspectives. First, people explore diverse application using GANs, such as image translation, portrait editing and video prediction. Second, some works focus on better generative structure for GANs, [DCGAN] use a deep convolution networks to generate real world images, [self attention GAN] involve attention into GANs which can improve the generative model's capability, [PGGAN] used a progressive growing strategy to achieve high resolution image. [StyleGAN] further improve the generated quality by using a new architecture. T hird, some other works focus on how to achieve better GANs by changing the GANs loss term. [Least-Squares GAN] use least-squares gan loss to reduce gradient vanishing problem, [WGAN] and [WGAN-GP] propose earth moving distance and gradient penalty to stable GANs training. In this paper, we propose a novel method which is orthogonal to all of these methods.

\noindent \textbf{Build Prior for Data Distribution}.
There are various methods that aim to build the prior for underlying data distribution. For example, Principle Component Analysis (PCA) ~\cite{turk1991face}, Independent Component Analysis (ICA)~\cite{hyvarinen2000independent}, and the Gaussian Mixture Model (GMM)~\cite{xu1996convergence, richardson2018gans}, all assume the data distribution follow a simple assumption. But they have difficulty modeling complex patterns of irregular distributions. Later works, such as the Hidden Markov Model (HMM)~\cite{starner1997real}, Markov Random Field (MRF)~\cite{ranzato2010generating}, and restricted Boltzmann machines (RBMs)~\cite{salakhutdinov2009deep}. limiting their results on texture patches, digital numbers, due to a lack of effective feature representations. Instead of building the model directly on the pixel space, we leverage the successful feature presentation of the CNNs~\cite{He_2016_CVPR,szegedy2016rethinking} and build the model on these feature representations.

%GANs aim to use a generative network to model target distribution mapping from a given latent space. There are still some method to model the data distribution. [on gans and gmms] propose to use Gaussian Mixture Model to model target distribution in pixel spaces, it's limited by the high dimensions. [GIQA] proposes to model real data distribution in feature space to help assess the generated image quality. In this paper, we also build a parametric model on feature space, using this prior to help training GANs.
\vspace{-0.3cm}
\section{Methods}

%In this paper, we pay main attention to the problem of \emph{low quality} and \emph{missing modes} in current GANs. These two problems are caused by the imprecise gradient of discriminator for generator. Suppose the discriminator function is $f(x)$, the gradient for generator is $\nabla_{x}f(x)$, the optimal discriminator is $f^{*}(x)$, then the gradient from optimal discriminator for generator is $\nabla_{x}f^{*}(x)$. 
%In Figure~\ref{fig:gan_problem}, we show that $\nabla_{x}f^{*}(x)$ fails to direct the generated samples which are out of real data distribution to the accurate direction. So the generator can't get a right update. 

%Unlike previous methods which attempt to apply regularization~\cite{zhou2019lipschitz,miyato2018spectral} or new loss functions~\cite{arjovsky2017wasserstein, mao2017least} to solve this problem. We propose real data prior to helping the discriminator solve the imprecise gradient problem. Equipped with the prior, we can solve the problem of \emph{low quality} and \emph{missing modes}. First, we can leverage the prior to enforce the poor generated samples to direct real data distribution. Second, we apply the prior to estimate the diversity of the generated samples, then we compare the diversity with the real data distribution. Then we can apply a resampling strategy to adjust the data distribution of a mini batch to adjust the diversity of the generated samples. Consequently, the \emph{missing modes} problem is been solved.

Before we introduce our methods, we first give a brief introduction and our findings about current GANs. In the vanilla GAN, the discriminator D and the generator G play the following two-player minimax game with value function $V(D, G)$:

\begin{equation}
\label{eq:minimaxgame-definition}
\min_{G} \max_D V(D, G) = \mathbb{E}_{x \sim p_{\text{r}}(x)}[\log D(x)] + \mathbb{E}_{x \sim G(z)}[\log (1 - D(x))].
\end{equation}

suppose $p_g(x) = G(z)$, for G fixed, the optimal discriminator $D$ for $V(D, G)$ is:
\begin{equation}
\label{eq:optimal-D}
   D^*_G(\bm{x}) = \frac{p_\text{r}(x)}{p_\text{r}(x) + p_g(x)}
\end{equation}

According to the analysis in GAN~\cite{goodfellow2014generative}, the global minimum of $V(D^*_G, G)$ is achieved if and only if $p_g = p_r$. But in the practical training process of GAN, the optimal discriminator $D_G^*(x)$ is usually hard to achieve due to insufficient training iterations or limited network capability. In this situation, the gradient is probably incorrect for reaching the optimal $p_g$.

%   notice: we discard this part for now !!! 
% Without the discriminator achieve $D_G^*(x)$, the solution for global minimum of $V(D^*_G, G)$ can not be achieved  when $p_g(x) = p_r(x)$.

% ~\textbf{Theorem 1.} If $D(x)$ is non-optimal: $D(x) \neq D^*_G(x)$, the global minimal of $V(D,G)$ could not be achieved when $p_g(x) = p_r(x)$.

% Due to the page limitation, the proof is in the supplementary material. ~\textbf{Theorem 1.} is easy to understand since the non-optimal discriminator offers the imprecise gradient to the generator and fails to guide the generated samples which are out of real data distribution to the accurate direction. So the generator can't get a right update.
%   notice: we discard this part for now !!!

\begin{figure*}[t]
\centering
 \includegraphics[width=1.0\columnwidth]{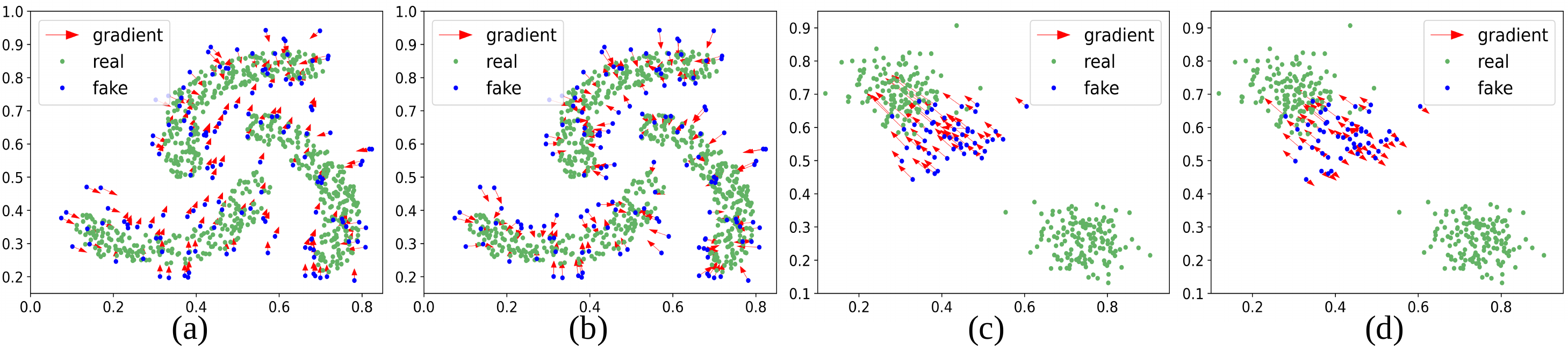}
 \vspace{-0.3cm}
\caption{An example of \emph{low quality} problem(a,b) and \emph{missing modes} problem(c,d) in GANs training on toy examples, (a) and (c) show the gradient of vanilla GAN, (b) and (d) show the gradient of our proposed PriorGAN.}
\label{fig:toy_vis}
\vspace{-0.3cm}
\end{figure*}

To demonstrate this, we train a basic GAN~\cite{goodfellow2014generative} on two typical toy distributions. As illustrated in Figure~\ref{fig:toy_vis}, The green points are the real data, the blue points are the generated data, the discriminator is trained based on current real and generated data distribution. The red arrows denote the gradient of the discriminator for each generated data. In Figure~\ref{fig:toy_vis}(a), some generated data are out of the real data distribution. But the gradients of these low quality generated points are not correct and they can't move to the real data distribution. So these samples stay in the situation of \emph{low quality} problem. In Figure~\ref{fig:toy_vis}(c), we notice that the real data distribution has two main regions, but the gradient of the generated points have the same direction to one region of the real data. This seems to be an intrinsic cause of the \emph{missing modes} problem.

To solve these problems, we propose to establish a prior model on real data to help the discriminator solve the inaccurate gradient problem. Equipped with the prior, we can solve the problem of \emph{low quality} and \emph{missing modes}. 
%First, we can leverage the prior to enforce the poor generated samples to direct real data distribution. Second, we apply the prior to estimate the diversity of the generated samples, then we compare the diversity with the real data distribution. Then we can apply a resampling strategy to adjust the data distribution of a mini batch to adjust the diversity of the generated samples. Consequently, the \emph{missing modes} problem is been solved.
In the next section, we will first introduce how to build real data prior. Then we present how to apply the prior as a loss term in GAN training to improve the quality of generated images. After that, we show how to use this prior to estimate the diversity of the real and generated data distribution. Finally, we introduce the resampling strategy to improve the diversity of the generated samples.

\subsection{Build Prior for Real Data}
%GANs aim to fit a distribution mapping function from source to target, and the target distribution was usually given by a set of images. Traditional GANs use sampling to get the character of real data distribution, but this sampling method may ignore some features according to Nyquist Sampling Law. So, we build a Gaussian Mixture Model to describe it. Considering the dimension of raw image was extremely high, we build this GMM model on deep feature domain. Suppose the feature extractor function is $f(\cdot)$, the mean vector and covariance matrix for $i$-th Gaussian component are $\bm{\mu}_i$ and $\mathbf{\Sigma}_i$, respectively. So the probability of an image $\mathcal{I}$ is given by:
Building an explicit prior to the high dimensional data is very difficult and challenging, especially for high-resolution images. This is because the real data distribution is extremely complicated to be captured by a simple prior. Previous method~\cite{richardson2018gans} try to leverage the Gaussian Mixture Model(GMM) model to build a model on the image patch level. However, the image resolution is restricted to be very small and the generated results are really blurry. Instead of directly building the GMM model on the pixel level, we take full advantage of the recent success of deep convolutional networks and build the GMM model on the features extracted by the networks. The deep feature is suitable for building the GMM model from two aspects. First, the deep feature is relatively low dimensional, which helps avoid the overfitting problem. Second, the deep feature is very representative and explicitly represents the content of the image. Denote the input image as $x$, the deep feature extractor function as $F(\cdot)$. We use a pre-trained network as feature extractor and the extracted feature dimension is $d$. Suppose the GMM model has $M$ Gaussian components, the mean vector and covariance matrix for $i$-th Gaussian component are $\mu_i$ and $\Sigma_i$, respectively. Given these parameters, the density function of a Gaussian Mixture Model takes the form
\begin{equation}
    p(x|\lambda) = \sum_{i=1}^{M} w_i \mathcal{N}(F(x)|\mu_i, \Sigma_i),   x \in P_r
    \label{eqn:GMM_probability}
\end{equation}

$P_r$ denotes real image space, $w_i$ is the mixture weights, which satisfy the constraint that $\sum_{i=1}^{M} w_i = 1$. And $\mathcal{N}(F(x)|\mu_i, \Sigma_i)$ is a $d$-variate Gaussian function of the form. The complete Gaussian mixture model is parameterized by the mean vectors, covariance matrices and mixture weights from all component densities. These parameters are collectively represented by the notation, $\lambda = \{w_i, \mu_i, \Sigma_i\}$. To estimate these parameters, we adopt the expectation-maximization (EM) algorithm~\cite{dempster1977maximum} to iteratively update them with the loss function
\begin{equation}
L= \sum_{x \in P_r} -\log p(x|\lambda).
\vspace{-0.2cm}
\end{equation}

\subsection{Improving Quality with Prior}
%The probability in prior can be seen as image quality[]. So a direct solution to attach high quality generated images is to maximize the image quality. We denote this loss term as $\mathcal{L}_{Q}$.
Given the estimated GMM model, for a generated image $x \in P_g$, we can adopt $p(\mathbf{x}|\lambda)$ in Equation.~\ref{eqn:GMM_probability} to estimate its probability. If its probability is high, it indicates that it is close to some of the real samples and the quality is high, otherwise, it is far from real samples and the quality is poor. So the generator should pay more attention to these poor quality samples. However, previous analysis shows that these poor quality generated samples may get inaccurate gradient from the discriminator. To solve this problem, we propose to utilize the prior to restrict these poor quality samples to get a higher probability in the built GMM model. This will help the poor quality samples to have an accurate gradient direction towards the real data distribution. Formally, given an generated image $x = G(z)$, $G$ is the generator, the loss for the generator is: 

\begin{equation}
\label{eqn:quality_score}
\mathcal{L}_{Q} = 
\begin{cases}
-\log p(F(G(z))| \lambda) & \textit{if } p(F(G(z))| \lambda) < \theta\\
0 & \textit{otherwise}
\end{cases},
\end{equation} 
where $\theta$ is the quality threshold. The quality loss $\mathcal{L}_{Q}$ mainly focuses on the low quality generated results and will give a right gradient direction for these low quality generated samples, as illustrated in Figure~\ref{fig:toy_vis}(b). During the training process of GANs, we give a loss weight $\delta$ to this quality loss and apply it as an auxiliary term to update the generator besides the GAN loss.

\subsection{Improving Diversity with Prior}
\begin{figure*}[t]
\centering
 \includegraphics[width=1.0\columnwidth]{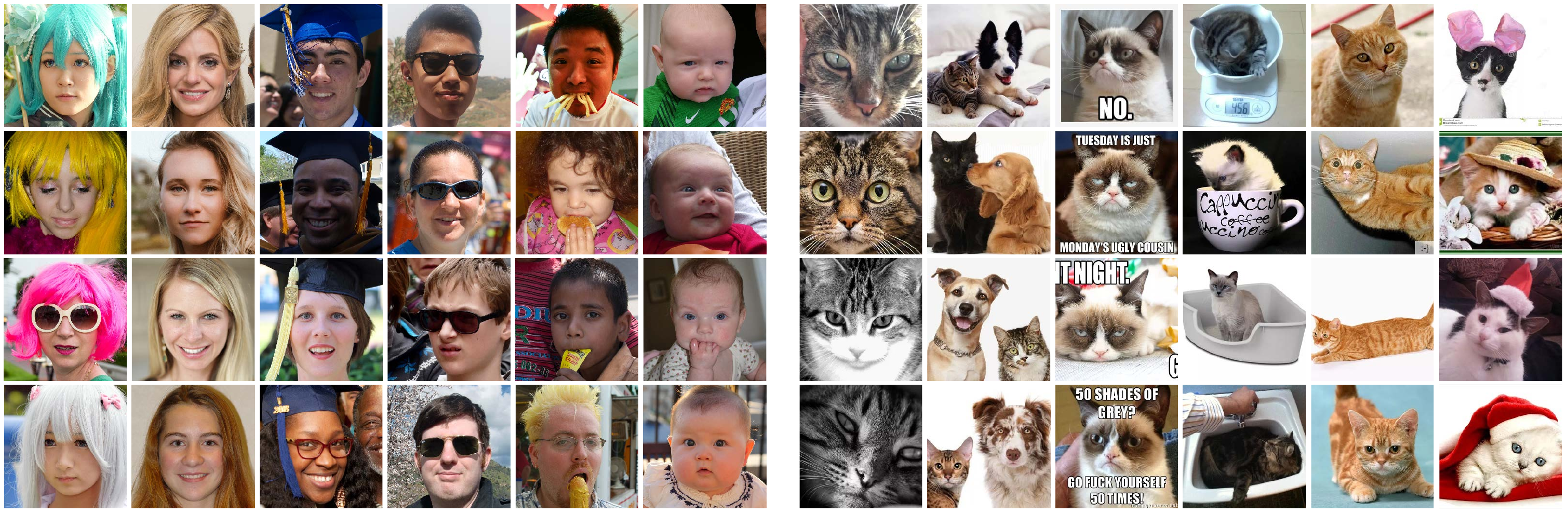}
 \vspace{-0.6cm}
\caption{Visualization of learned GMM-based prior on FFHQ and LSUN-cat dataset, each column are images randomly selected from the same Gaussian component. We can observe that images within the same Gaussian component share the same attributes, like hat, hair, sunglasses for the FFHQ or eyes, color, posture for the LSUN-cat.}
\label{fig:prior}
\vspace{-0.2cm}
\end{figure*}

Current GANs often suffer from the \emph{missing modes} problem, the learned distribution from GANs may miss some certain regions of the real data distribution. Therefore the key is how we can estimate the diversity of real data distribution and find these missing regions in the learned distribution. One possible solution is to apply the built GMM model. Specifically, if we establish a GMM model to represent a real data distribution, then we can apply the $M$ Gaussian components in the GMM model to divide the real data distribution into $M$ groups and estimate the frequency of the real samples in each group. A sample belonging to which Gaussian component is decided by the minimum distance to the center of each Gaussian component, as we shown in Figure~\ref{fig:prior}. Suppose the number of the real samples on the $i$-th Gaussian component is $n_i^r$, then the frequency of the real samples on the $i$-th Gaussian component is $f_i^r = {n_i^r}/{\sum_i^M{n_i^r}}$.  And we have the vector $[f_1^r, f_2^r,\cdots, f_M^r]$ to represent the diversity of the real samples. Similarly, we apply the same GMM model to get the frequency of generated samples, which is denoted as $[f_1^g, f_2^g,\cdots, f_M^g]$. 

\begin{figure*}[t]
\centering
 \includegraphics[width=1.0\columnwidth]{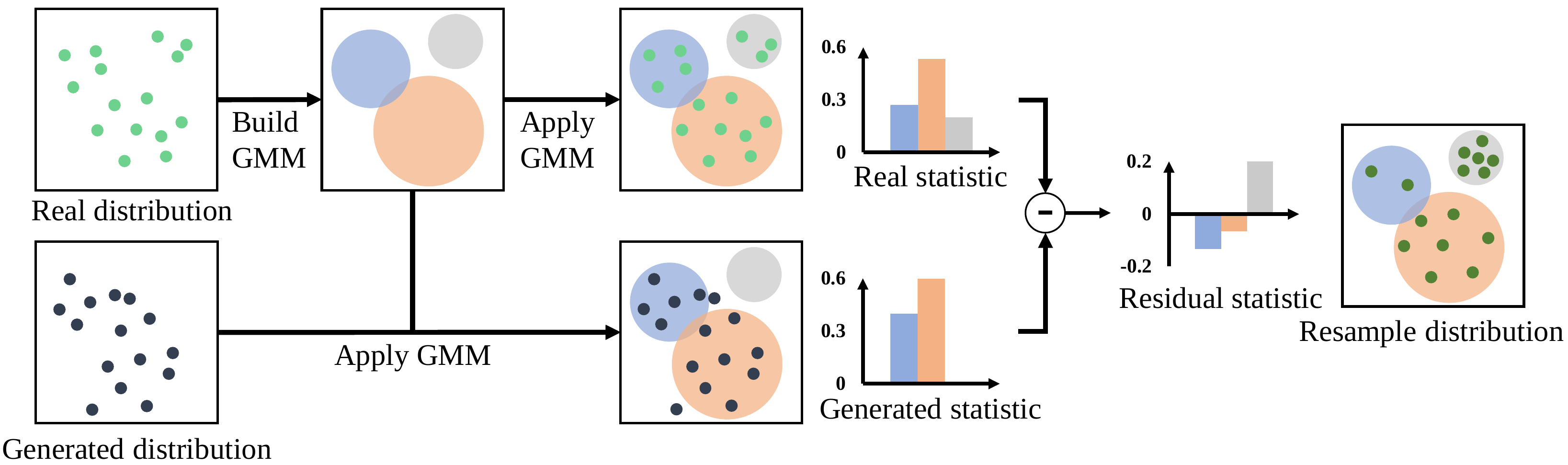}
 \vspace{-0.7cm}
 \caption{illustration of the pipeline of our resampling strategy.}
\label{fig:resample}
\vspace{-0.4cm}
\end{figure*}

The frequency difference between the real and generated samples can indicate whether the missing modes problem occurs. If the learned distribution misses a certain region $S$ of the real data distribution, and $S$ belongs to the $i$-th Gaussian component of the built GMM model, then $i$-th learned data frequency $f_i^g$ will be smaller than the real data frequency $f_i^r$. So the diversity difference between real data distribution and learned distribution can be represented by the frequency distances. Suppose the diversity distance between real data distribution $p_r$ and learned distribution $p_g$ is denoted as  
\begin{equation}
    d(p_r,p_g) = [f_1^r, f_2^r,\cdots, f_M^r] - [f_1^g, f_2^g,\cdots, f_M^g].
\end{equation}
So $d(p_r, p_g)$ is a vector with $M$ elements, the $i$-th element is $d_i(p_r,p_g)$. So the overall diversity differences score (DDS) is 
\begin{equation}
DDS(p_r, p_g) = \sum_i^M |d_i(p_r,p_g)|. 
\end{equation}

An ideal learned distribution should satisfy $DDS(p_r, p_g)=0$. However, the $DDS$ is non-differentiable and cannot be directly used as a loss function. So we optimize it in an indirect way and proposed a resampling strategy. Considering a typical missing modes phenomenon where the real data distribution has two regions $S_1$ and $S_2$, the generated samples only lies around the $S_1$, current discriminator can't force the generated samples around $S_1$ to move to $S_2$ since the discriminator consider the generated samples around $S_1$ are "real enough". To overcome this problem, we propose to change the decision boundary of the discriminator. If we sample fewer samples in $S_1$ and more samples in $S_2$ for training the discriminator, the decision boundary of the discriminator will be changed. The discriminator will encourage the generated samples to move to $S_2$ and solve the \emph{missing modes} problem, as illustrated in Figure~\ref{fig:toy_vis}(d). We call this process the resampling strategy. This strategy will change the real data distribution in the training procedure to help the generator overcome the missing modes problem. Formally, we use a new sampling frequency $f^{r'}_i$ to sample the real samples in the $i$-th Gaussian component for training. $f^{r'}_i$ is decided by the original frequency and the diversity distances:
\begin{equation}
    f^{r'}_i = max(f_{i}^{r} + \alpha d_i(p_r,p_g), 0),
\end{equation}
Then we normalized the sampling frequency, $f^{r'}_i = {f^{r'}_i}/ {\sum_i^M f^{r'}_i}$. Such that the new sampling frequency is between $0$ and $1$, $\alpha$ is a hyper parameter which decides the resample weight. For $f_g^i < f_r^i$, we increase the sampling frequency of region $i$ and decrease it if otherwise. The resampling strategy is also illustrated in Figure~\ref{fig:resample}.

Above we introduce our methods for improving quality and diversity with our proposed prior. This prior can be integrated into various frameworks of GANs, such as WGAN-GP~\cite{gulrajani2017improved}, LSGAN~\cite{mao2017least}, SNGAN~\cite{miyato2018spectral}, and even the StyleGAN~\cite{karras2019style}. This flexible combination with existing methods can help these methods get further improvement.

\section{Experiments}

%\subsection{Implementation details}
In this part, we conduct experiments on a large variety of generative models trained on different datasets including CIFAR-10~\cite{krizhevsky2009learning}, LSUN~\cite{yu2015lsun}, and FFHQ~\cite{karras2019style} datasets to evaluate the performance of our proposed methods.

\textbf{Datasets.}
CIFAR-10 is a widely used dataset for image generation. We directly apply the original image size $32 \times 32$ for training. LSUN~\cite{yu2015lsun} is widely used for high-resolution image generation, we choose two categories in LSUN for training and evaluation: cat and bird, which we call LSUN-cat and LSUN-bird, respectively. LSUN-cat contains $1.6$ million images and LSUN-bird contains $2.2$ million images, we follow the settings in StyleGAN~\cite{karras2019style} to resize all the images to $256 \times 256$ resolution. FFHQ~\cite{karras2019style} is a recently high-resolution face dataset contains $70000$ images, we resize them to $256 \times 256$ resolution for training, which we call FFHQ-256.

\textbf{Evaluations.}
To evaluate the quality of an image set, as we mentioned earlier, the probability of each image in our GMM model can indicate its quality. So we use the mean of normalized image density in Equation~\ref{eqn:GMM_probability} to evaluate the quality of an image set, calculated as:  $ \frac{1}{K} \sum_{\mathbf{i}=1}^K norm( log ( p(x_i|\lambda)))$, $K$ is the number of images in the image set\footnote[1]{More implementation details please refer to supplementary material.}. We named it Quality Score(QS), higher QS indicates better image quality. And we use our proposed DDS to evaluate the diversity of the generative model, FID~\cite{heusel2017gans} to measure both quality and diversity. 
%FID is a commonly used evaluation methods for diversity and quality evaluation. QS and DDS can independently evaluate the quality and diversity of each generative model.
For a fair comparison, we would use $5k$ or $50k$ genearted images to calculate the FID, denoted as FID(5K) and FID(50K) respectively. $10k$ images are used to calculate QS and DDS for the CIFAR-10 dataset. For the FFHQ, LSUN-cat, and LSUN-bird datasets, we use $50k$ images to calculate the result of the FID, QS and DDS.

\textbf{Implementation Details.}
For each baseline, we follow training settings in their experiments. We only add quality loss and the resampling strategy during the training of each baseline. In order to build the prior for each dataset, we apply ResNet101~\cite{He_2016_CVPR} or InceptionV3~\cite{szegedy2016rethinking} to extract the features for all the real images. The features we use are from the last average pooling layer. Without special notes, we use ResNet101 to extract the features. 
Then we build the prior for each dataset. Specifically, we set the number of Gaussian component to $35$,$70$, $7$, and $7$, the resample weight $\alpha$ to $3$,$7$,$7$,$7$ for the CIFAR-10, FFHQ, LSUN-cat, and LSUN-bird, respectively, and the quality loss weight $\delta$ set to $0.1$. More details could be found in the supplementary material.

% training and testing detail

\begin{table}[t]
\centering
  \small{
  \begin{tabular}[t]{c|ccc}
    % \toprule
    \hline
    Method & Inception score $\uparrow$ & FID(5k)$\downarrow$ &  FID(50k)$\downarrow$ \\
    % \midrule
    \shline
    Real data & 11.24$\pm$.12  & 8.63$\pm$.0 & - \\
    % \midrule
    \hline
    DCGAN~\cite{radford2015unsupervised}& 6.64$\pm$.14 & - & - \\
    LR-GANs~\cite{yang2017lr} &  7.17$\pm$.07 & - & - \\
    Warde-Farley et al.~\cite{warde2016improving} & 7.72$\pm$.13 & - & - \\
    WGAN-GP~\cite{gulrajani2017improved} & 7.86$\pm$.08 & 40.2$\pm$.0 & - \\
    %SNGAN~\cite{miyato2018spectral} & 8.22$\pm$.05 & 21.7$\pm$.21  \\
    %PriorSNGAN(ours) & 8.60$\pm$.19 & 18.19$\pm$.06  \\
    LGAN~\cite{zhou2019lipschitz} & 8.03$\pm$.03 & - & 15.64$\pm$.07 \\
    SNGAN~\cite{miyato2018spectral} & 8.22$\pm$.05 & 21.7$\pm$.21 & 13.78$\pm$.11  \\
    $L^{10}$ BWGAN~\cite{adler2018banach} & 8.31$\pm$.07 & - & -  \\
    QAGAN~\cite{NIPS2019_8560} & 8.37$\pm$.04 & - & 13.91$\pm$.11  \\
    WGAN-ALP~\cite{terjek2019virtual} & 8.34$\pm$.06 & - & 12.96$\pm$.35 \\
    %AntoGAN~\cite{gong2019autogan}$(50k)$ & 8.55$\pm$.10 & 12.42$\pm$.00 \\
    PriorSNGAN(\textbf{ours}) & \textbf{8.60$\pm$.19} & \textbf{18.19$\pm$.06} & \textbf{ 11.94$\pm$.18}   \\
    % \bottomrule
    \shline
  \end{tabular}
  \vspace{0.1cm}
    \caption{\label{tab:incepscores}Inception scores and FIDs with unconditional image generation on CIFAR-10. FID$(50k)$ means calculating the score with $50k$ generated images, otherwise with $5k$ generated images.
  }
  \vspace{-0.5cm}
  }
\end{table}

\begin{comment}
\begin{table}[t]
\centering
  \small{
  \begin{tabular}[t]{lcc}
    \toprule
    Method & Inception score $\uparrow$  &  FID(50k)$\downarrow$ \\
   
    \midrule
    Real data & 11.24$\pm$.12  &  \\
    \midrule
    DCGAN~\cite{radford2015unsupervised}& 6.64$\pm$.14 & \\
    LR-GANs~\cite{yang2017lr} &  7.17$\pm$.07\\
    Warde-Farley et al.~\cite{warde2016improving} & 7.72$\pm$.13 &  \\
    WGAN-GP~\cite{gulrajani2017improved} & 7.86$\pm$.08 & 19.65 \\
    %SNGAN~\cite{miyato2018spectral} & 8.22$\pm$.05 & 21.7$\pm$.21  \\
    %PriorSNGAN(ours) & 8.60$\pm$.19 & 18.19$\pm$.06  \\
    LGAN~\cite{zhou2019lipschitz} & 8.03$\pm$.03 & 15.64$\pm$.07 \\
    SNGAN~\cite{miyato2018spectral} & 8.22$\pm$.05 & 13.78$\pm$.11  \\
    $L^{10}$ BWGAN~\cite{adler2018banach} & 8.31$\pm$.07 & -  \\
    WGAN-ALP~\cite{terjek2019virtual} & 8.34$\pm$.06 & 12.96$\pm$.35 \\
    %AntoGAN~\cite{gong2019autogan}$(50k)$ & 8.55$\pm$.10 & 12.42$\pm$.00 \\
    PriorSNGAN(\textbf{ours}) & \textbf{8.60$\pm$.19} & \textbf{ 11.94$\pm$.18}   \\
    \bottomrule
  \end{tabular}
  \vspace{0.1cm}
    \caption{\label{tab:incepscores}Inception scores and FIDs with unconditional image generation on CIFAR-10. FID$(50k)$ means calculating the score with $50k$ images otherwise with $5k$ generated images.
  }
  }
\end{table}
\end{comment}

\subsection{Comparison with state-of-the-art Methods}
In this section, we compare our results with other state-of-the-art methods. We conduct the experiments on the CIFAR-10 dataset. The basic framework is an unconditional GAN, the generator and discriminator network structure are the same as the ResNet structure adopted in the baseline SNGAN~\cite{miyato2018spectral}. We use $5k$ randomly generated images to calculate the Inception score ten times and report the mean and variance. For the FID, there are two different settings to calculate the result, some methods~\cite{miyato2018spectral, heusel2017gans} use $5k$ generated images and $10k$ real images to calculate, while the others~\cite{parimala2019quality,terjek2019virtual} use $50k$ generated and $50k$ real images to calculate. Using $50k$ generated images can get a better result. For a fair comparison, we report both results and denote them as the FID($5k$) and the FID($50k$). As shown in Table~\ref{tab:incepscores}, our method (PriorSNGAN) outperforms the baseline SNGAN and other state-of-the-art GANs. These results confirm the effectiveness of our proposed real data prior to GAN training.
% without any bells and whistles

\begin{table}[t]
\centering
  \begin{tabular}{c|ccc|ccc|ccc}
    \hline
    \multirow{2}{*}{ } &
      \multicolumn{3}{c|}{SNGAN } &   
      \multicolumn{3}{c|}{WGAN-GP } &  
      \multicolumn{3}{c}{LSGAN } \\
    \cline{2-10}
    &  FID$\downarrow$ &  QS$\uparrow$  & DDS$\downarrow$ & FID$\downarrow$& QS$\uparrow$ & DDS$\downarrow$ &  FID$\downarrow$ &  QS$\uparrow$  & DDS$\downarrow$   \\
    \shline
    Baseline & 13.78 & 0.429  & 0.488 & 19.65 & 0.410  & 0.532 & 34.90 & 0.367  & 0.545 \\
    % \hline
    +Prior & 11.94 & 0.512  & 0.380 & 18.16 & 0.439  & 0.475 & 30.24 & 0.406 & 0.487 \\
    \shline
  \end{tabular}
  \vspace{0.14cm}
\caption{Comparison of FID, QS, and DDS metrics on three different GANs: SNGAN, WGAN-GP and LSGAN.}
\label{table:model_score}
\vspace{-0.5cm}
\end{table}

\subsection{Prior for Various GANs}
Our approach enjoys a high degree of flexibility and can be integrated into various kinds of GAN frameworks. We choose some variants of GAN as baseline frameworks, such as LSGAN~\cite{mao2017least}, WGAN-GP~\cite{gulrajani2017improved}, SNGAN~\cite{miyato2018spectral}, and then integrate our method into these frameworks for comparison. For a fair comparison, we follow the same settings and architecture of GANs and only apply our quality loss and the resampling strategy, which are denoted as +Prior. 

We conducted experiments on the CIFAR-10 dataset, Table~\ref{table:model_score} reports a comparison of our method with baseline models. We can observe that our methods outperform the corresponding baseline GAN frameworks in terms of FID, QS, and DDS metrics, which demonstrate the superiority of our proposed framework.

%Our methods achieve the best results not only on quality but also on diversity.

\begin{figure*}[t]
\centering
 \includegraphics[width=1.0\columnwidth]{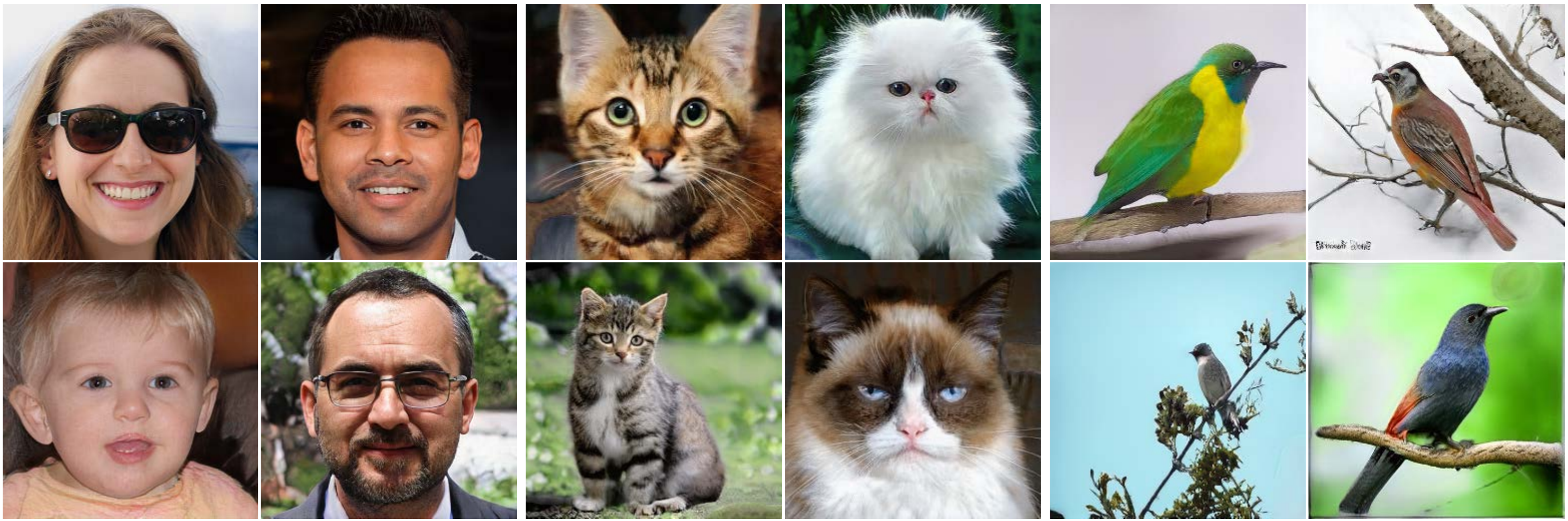}
 \vspace{-0.4cm}
\caption{Visualization results of our generated images on FFHQ-256, LSUN-cat and LSUN-bird datasets, our method can generate not only high quality but also diverse images.}
\label{fig:vis_results}
\vspace{-0.1cm}
\end{figure*}

\begin{table}[t]
%\scriptsize
%\footnotesize
\centering
  \begin{tabular}{c|ccc|ccc|ccc}
    \hline
    \multirow{2}{*}{ } &
      \multicolumn{3}{c|}{FFHQ-256 } &   
      \multicolumn{3}{c|}{LSUN-cat } &   
      \multicolumn{3}{c}{LSUN-bird } \\%&  
      %\multicolumn{3}{c}{FFHQ-1024 } \\
    \cline{2-10}
    &  FID$\downarrow$ &  QS$\uparrow$  & DDS$\downarrow$ & FID$\downarrow$& QS$\uparrow$ & DDS$\downarrow$ &  FID$\downarrow$ &  QS$\uparrow$  & DDS$\downarrow$   \\
    % \hline
    \shline
    StyleGAN & 11.69 & 0.703 & 0.250 & 9.66 & 0.486 & 0.203 & 10.26 & 0.513 & 0.136 \\
    % \hline
    +Prior(R101) & 8.79 & 0.724 & 0.199 & 8.25 & 0.526 & 0.162 & 9.04 & 0.530 & 0.109 \\
    % \hline
    +Prior(InV3)& 7.12 & 0.749 & 0.170 & 7.21 & 0.553 & 0.151 & 8.44 & 0.554 & 0.086 \\
    \shline
  \end{tabular}
  \vspace{0.2cm}
\caption{Comparison of FID, QS, and DDS metrics on three different datasets: FFHQ-256, LSUN-cat, and LSUN-bird.}
\label{table:stylegan_score}
\vspace{-0.2cm}
\end{table}

\begin{table}[t]
% \scriptsize
%\footnotesize
\centering
  \begin{tabular}{c|ccc|ccc|ccc}%|ccc}
    \hline
    \multirow{2}{*}{ } &
      \multicolumn{3}{c|}{FFHQ-256 } &   
      \multicolumn{3}{c|}{LSUN-cat } &   
      \multicolumn{3}{c}{LSUN-bird } \\%&  
      %\multicolumn{3}{c}{FFHQ-1024 } \\
    \cline{2-10}
    &  FID$\downarrow$ &  QS$\uparrow$  & DDS$\downarrow$ & FID$\downarrow$& QS$\uparrow$ & DDS$\downarrow$ &  FID$\downarrow$ &  QS$\uparrow$  & DDS$\downarrow$ \\ % & FID$\downarrow$& QS$\uparrow$ & DS$\uparrow$   \\
    % \hline
    \shline
    StyleGAN & 11.69 & 0.703 & 0.250 & 9.66 & 0.486 & 0.203 & 10.26 & 0.513 & 0.136 \\% & 4.44 & 0.658 & 0.582 \\
    % \hline
    StyleGAN +$\mathcal{L}_Q$ & 9.75 & 0.732 & 0.229 & 8.85 & 0.531 & 0.188 & 9.17 & 0.534 & 0.123 \\ %& 4.15 & 0.677 & 0.579 \\
    % \hline
    StyleGAN +RS & 10.00 & 0.715 & 0.196 & 9.11 & 0.505 & 0.149 & 9.79 & 0.521  & 0.095 \\% & 4.33 & 0.660 & 0.584 \\
    % \hline
    %Ours(ResNets101) & 8.79 & 0.724 & 0.626 & 8.25 & 0.526 & 0.446 & 9.04 & 0.530 & 0.407 & 4.10 & 0.671 & 0.584 \\
    % \hline
    %StyleGAN+OHEM~\cite{gu2020giqa} & 11.20 & 0.715 & 0.623 & 9.42 & 0.512 & 0.444 & 9.83 & 0.525 & 0.405 & 4.31 & 0.664 & 0.583 \\
    % \hline
    %Ours(InceptionV3) & 7.12 & 0.749 & 0.628 & 7.20 & 0.553 & 0.449 & 8.44 & 0.554 & 0.407 & 4.04 & 0.673 & 0.585 \\
    \shline
  \end{tabular}
  \vspace{0.14cm}
\caption{Ablation study of our proposed prior with only quality loss and only resampling strategy.}
\label{table:ablation_study}
\vspace{-0.25cm}
\end{table}

\subsection{Prior for Various Datasets}
We also try to apply our proposed prior for various datasets, such as FFHQ-256, LSUN-cat and LSUN-bird. StyleGAN~\cite{karras2019style} recently achieve great success for high-resolution image generation. In this section, we try to combine our method with StyleGAN and evaluate the results on these datasets. We apply two different pre-trained models, ResNet101~\cite{He_2016_CVPR} and InceptionV3~\cite{szegedy2016rethinking} to extract deep features for building the GMM model. We call them +Prior(R101) and +Prior(InV3), respectively. The quantitative results are shown in Table~\ref{table:stylegan_score}. We can observe that our approach can effectively improve the quality(QS), diversity(DDS), and the overall score(FID) over the strong baseline StyleGAN by large margins on all datasets. And the real data prior build on the feature extracted with InceptionV3 model achieves better results than ResNet101, one possible reason is that the evaluation methods are also based on InceptionV3 model. Some visualization results are shown in Figure~\ref{fig:vis_results}.

% \subsection{Visualization of the Learned Prior}
% To understand what each Gaussian component in a GMM model learns for real data distribution. We provide the visualization results of each Gaussian component in the learned GMM model. We randomly choose some Gaussian components from a GMM model and show some images from these components. Figure~\ref{fig:prior} shows the results on the GMM model built with the FFHQ and LSUN-cat dataset. The images in each column belong to the same Gaussian component in the built GMM model. We can observe that images within the same Gaussian component share the same attributes, like hat, makeup, sunglasses on the face or breed, color, posture of the cat. This also proves that these Gaussian components can represent multiple modes of the real data and help to solve the \emph{missing modes} problem. More results are in the supplementary material.
%We strongly recommend readers read our supplemental material for more impressive clustering results and implementation details.

\begin{table}[h!]
    \begin{minipage}{\columnwidth}
		\begin{minipage}[t]{0.3\columnwidth}
			\small
			\centering
			\begin{tabular}{c|ccc}
			    \hline
				$M$  &  FID$\downarrow$  & QS$\uparrow$ & DDS$\downarrow$ \\ 
				\shline
				 5 & 7.26 & 0.548 & 0.156 \\
				 % \hline
				 7 & 7.21  & 0.553 & 0.151 \\
				 % \hline
				 10 & 7.32  & 0.552 & 0.159 \\
				 % \hline
				 20 & 7.36  & 0.549 & 0.163 \\
				 % \hline
				 50 & 7.45  & 0.543 & 0.177 \\
				 \shline
			\end{tabular}
			\vspace{0.08cm}
			\makeatletter\def\@captype{table}\makeatother\caption{PriorGAN results on LSUN-cat dataset with different number of Gaussian components $M$.}
			\label{table:kernel-densities}
			\vspace{-0.3cm}
		\end{minipage}
		\quad 
		\begin{minipage}[t]{0.3\columnwidth}
			\small
			\centering
			\begin{tabular}{c|ccc}
			    \hline
				$\delta$  &  FID$\downarrow$  & QS$\uparrow$ & DDS$\downarrow$ \\ 
				\shline
				 0.03 & 8.31  & 0.509 & 0.191 \\
				 % \hline
				 0.07 & 7.82  & 0.529 & 0.201 \\
				 % \hline
				 0.10 & 7.66  & 0.550 & 0.195 \\
				 % \hline
				 0.15 & 8.11  & 0.578 & 0.220 \\
				 % \hline
				 0.2 & 8.29  & 0.586 & 0.227 \\
				 \shline
			\end{tabular}
			\vspace{0.08cm}
			\makeatletter\def\@captype{table}\makeatother\caption{PriorGAN  results on LSUN-cat dataset with different quality loss weight $\delta$.}
			\label{table:quality-loss-weight}
			\vspace{-0.3cm}
		\end{minipage}
		\quad
		\begin{minipage}[t]{0.3\columnwidth}
			\small
			\centering
			\begin{tabular}{c|ccc}
			    \hline
				$\alpha$  &  FID$\downarrow$  & QS$\uparrow$ & DDS$\downarrow$ \\ 
				\shline
				 1 & 8.76  & 0.499  & 0.177 \\
				% \hline
				 3 & 8.35 & 0.506 & 0.162 \\
				% \hline
				 5 & 8.30 & 0.514 & 0.143 \\
				% \hline
				 7 & 8.08  & 0.521 & 0.149 \\
				% \hline
				 9 & 8.16 & 0.515 & 0.145 \\
				\shline
			\end{tabular}
			\vspace{0.2cm}
			\makeatletter\def\@captype{table}\makeatother\caption{PriorGAN results on LSUN-cat dataset with different resample weight $\alpha$. }
			\label{table:resample-weight}
			\vspace{-0.3cm}
		\end{minipage}
	\end{minipage}
\end{table}

\subsection{Ablation Study}
To validate the effectiveness of our proposed two key methods: the quality loss and resampling strategy, we apply additional experiments on FFHQ-256, LSUN-cat, LSUN-bird dataset. The quality loss $\mathcal{L}_Q$ and resampling strategy are applied independently to StyleGAN. The resulting models are called StyleGAN +$\mathcal{L}_Q$ and StyleGAN +RS. The results are shown in Table~\ref{table:ablation_study}. We can observe that adding the loss function $\mathcal{L}_Q$ mainly improve QS, which indicates the loss function $\mathcal{L}_Q$ mainly solve the \emph{low quality} problem. Adding the resampling strategy can get a lower DDS, which indicates that the resampling strategy mainly solves the \emph{missing modes} problem. This further validates the effectiveness of our proposed methods.

\subsection{Hyper Parameter Analysis}
In this section, we conduct experiments to investigate the sensitiveness of the number of Gaussian components $M$, quality loss weight $\delta$, and resample weight $\alpha$ in our proposed method. All the experiments are applied to the LSUN-cat dataset.

To reduce the computation cost, we apply PCA on the feature extracted from InceptionV3 model and keep 98\% of the total variance by default. First, we explore how the number of Gaussian components $M$ affects the results. We set $M$ to 5,7,10,20,50 and test the results on LSUN-cat dataset, we show the results in Table~\ref{table:kernel-densities}. The results are not very sensitive to $M$ and we achieve the best performance when setting it to 7. Besides, we set $M$ to 7 and explore how quality loss weight $\delta$ and resample weight $\alpha$ affects the performance, the results are shown in Table ~\ref{table:quality-loss-weight},~\ref{table:resample-weight}.

\section{Conclusion}
In this paper, we analyze current GANs suffer from low quality and missing modes problem. To tackle this problem, we propose a novel method which builds a GMM prior on deep feature space, by using this GMM prior, we propose quality loss to increase generated image quality and resampling strategy to enrich the diversity. Experiments have shown our method can easily insert into any GAN baseline and outperforms state-of-the-art by large margins.

\section*{Broader Impact}
Our PriorGAN is a general method for image generation. From the positive aspect, many generation tasks like image-to-image translation, image editing will benefit from it. We hope our work will serve as a solid baseline and can ease future research in generative models. From the negative aspect, our method can also be applied to some image manipulation techniques, which may cause severe trust issues and security concerns in our society.

{\small
\bibliographystyle{ieee_fullname}
\bibliography{egbib}

\begin{thebibliography}{10}\itemsep=-1pt

\bibitem{adler2018banach}
Jonas Adler and Sebastian Lunz.
\newblock Banach wasserstein gan.
\newblock In {\em Advances in Neural Information Processing Systems}, pages
  6754--6763, 2018.

\bibitem{arjovsky2017wasserstein}
Martin Arjovsky, Soumith Chintala, and L{\'e}on Bottou.
\newblock Wasserstein gan.
\newblock {\em arXiv preprint arXiv:1701.07875}, 2017.

\bibitem{bao2017cvae}
Jianmin Bao, Dong Chen, Fang Wen, Houqiang Li, and Gang Hua.
\newblock Cvae-gan: fine-grained image generation through asymmetric training.
\newblock In {\em Proceedings of the IEEE International Conference on Computer
  Vision}, pages 2745--2754, 2017.

\bibitem{brock2018large}
Andrew Brock, Jeff Donahue, and Karen Simonyan.
\newblock Large scale gan training for high fidelity natural image synthesis.
\newblock {\em arXiv preprint arXiv:1809.11096}, 2018.

\bibitem{cao2019multi}
Jiezhang Cao, Langyuan Mo, Yifan Zhang, Kui Jia, Chunhua Shen, and Mingkui Tan.
\newblock Multi-marginal wasserstein gan.
\newblock In {\em Advances in Neural Information Processing Systems}, pages
  1774--1784, 2019.

\bibitem{dempster1977maximum}
Arthur~P Dempster, Nan~M Laird, and Donald~B Rubin.
\newblock Maximum likelihood from incomplete data via the em algorithm.
\newblock {\em Journal of the Royal Statistical Society: Series B
  (Methodological)}, 39(1):1--22, 1977.

\bibitem{dong2019margingan}
Jinhao Dong and Tong Lin.
\newblock Margingan: Adversarial training in semi-supervised learning.
\newblock In {\em Advances in Neural Information Processing Systems}, pages
  10440--10449, 2019.

\bibitem{goodfellow2014generative}
Ian Goodfellow, Jean Pouget-Abadie, Mehdi Mirza, Bing Xu, David Warde-Farley,
  Sherjil Ozair, Aaron Courville, and Yoshua Bengio.
\newblock Generative adversarial nets.
\newblock In {\em Advances in neural information processing systems}, pages
  2672--2680, 2014.

\bibitem{gu2019mask}
Shuyang Gu, Jianmin Bao, Hao Yang, Dong Chen, Fang Wen, and Lu Yuan.
\newblock Mask-guided portrait editing with conditional gans.
\newblock In {\em Proceedings of the IEEE Conference on Computer Vision and
  Pattern Recognition}, pages 3436--3445, 2019.

\bibitem{gulrajani2017improved}
Ishaan Gulrajani, Faruk Ahmed, Martin Arjovsky, Vincent Dumoulin, and Aaron~C
  Courville.
\newblock Improved training of wasserstein gans.
\newblock In {\em Advances in neural information processing systems}, pages
  5767--5777, 2017.

\bibitem{He_2016_CVPR}
Kaiming He, Xiangyu Zhang, Shaoqing Ren, and Jian Sun.
\newblock Deep residual learning for image recognition.
\newblock In {\em The IEEE Conference on Computer Vision and Pattern
  Recognition (CVPR)}, June 2016.

\bibitem{heusel2017gans}
Martin Heusel, Hubert Ramsauer, Thomas Unterthiner, Bernhard Nessler, and Sepp
  Hochreiter.
\newblock Gans trained by a two time-scale update rule converge to a local nash
  equilibrium.
\newblock In {\em Advances in neural information processing systems}, pages
  6626--6637, 2017.

\bibitem{hyvarinen2000independent}
Aapo Hyv{\"a}rinen and Erkki Oja.
\newblock Independent component analysis: algorithms and applications.
\newblock {\em Neural networks}, 13(4-5):411--430, 2000.

\bibitem{isola2017image}
Phillip Isola, Jun-Yan Zhu, Tinghui Zhou, and Alexei~A Efros.
\newblock Image-to-image translation with conditional adversarial networks.
\newblock In {\em Proceedings of the IEEE conference on computer vision and
  pattern recognition}, pages 1125--1134, 2017.

\bibitem{karras2017progressive}
Tero Karras, Timo Aila, Samuli Laine, and Jaakko Lehtinen.
\newblock Progressive growing of gans for improved quality, stability, and
  variation.
\newblock {\em arXiv preprint arXiv:1710.10196}, 2017.

\bibitem{karras2019style}
Tero Karras, Samuli Laine, and Timo Aila.
\newblock A style-based generator architecture for generative adversarial
  networks.
\newblock In {\em Proceedings of the IEEE Conference on Computer Vision and
  Pattern Recognition}, pages 4401--4410, 2019.

\bibitem{krizhevsky2009learning}
Alex Krizhevsky, Geoffrey Hinton, et~al.
\newblock Learning multiple layers of features from tiny images.
\newblock 2009.

\bibitem{mao2019mode}
Qi Mao, Hsin-Ying Lee, Hung-Yu Tseng, Siwei Ma, and Ming-Hsuan Yang.
\newblock Mode seeking generative adversarial networks for diverse image
  synthesis.
\newblock In {\em Proceedings of the IEEE Conference on Computer Vision and
  Pattern Recognition}, pages 1429--1437, 2019.

\bibitem{mao2017least}
Xudong Mao, Qing Li, Haoran Xie, Raymond~YK Lau, Zhen Wang, and Stephen
  Paul~Smolley.
\newblock Least squares generative adversarial networks.
\newblock In {\em Proceedings of the IEEE International Conference on Computer
  Vision}, pages 2794--2802, 2017.

\bibitem{miyato2018spectral}
Takeru Miyato, Toshiki Kataoka, Masanori Koyama, and Yuichi Yoshida.
\newblock Spectral normalization for generative adversarial networks.
\newblock {\em arXiv preprint arXiv:1802.05957}, 2018.

\bibitem{NIPS2019_8560}
KANCHARLA PARIMALA and Sumohana Channappayya.
\newblock Quality aware generative adversarial networks.
\newblock In {\em Advances in Neural Information Processing Systems 32}, pages
  2948--2958. Curran Associates, Inc., 2019.

\bibitem{parimala2019quality}
KANCHARLA PARIMALA and Sumohana Channappayya.
\newblock Quality aware generative adversarial networks.
\newblock In {\em Advances in Neural Information Processing Systems}, pages
  2944--2954, 2019.

\bibitem{perarnau2016invertible}
Guim Perarnau, Joost Van De~Weijer, Bogdan Raducanu, and Jose~M {\'A}lvarez.
\newblock Invertible conditional gans for image editing.
\newblock {\em arXiv preprint arXiv:1611.06355}, 2016.

\bibitem{radford2015unsupervised}
Alec Radford, Luke Metz, and Soumith Chintala.
\newblock Unsupervised representation learning with deep convolutional
  generative adversarial networks.
\newblock {\em arXiv preprint arXiv:1511.06434}, 2015.

\bibitem{ranzato2010generating}
Marc'aurelio Ranzato, Volodymyr Mnih, and Geoffrey~E Hinton.
\newblock Generating more realistic images using gated mrf's.
\newblock In {\em Advances in Neural Information Processing Systems}, pages
  2002--2010, 2010.

\bibitem{richardson2018gans}
Eitan Richardson and Yair Weiss.
\newblock On gans and gmms.
\newblock In {\em Advances in Neural Information Processing Systems}, pages
  5847--5858, 2018.

\bibitem{salakhutdinov2009deep}
Ruslan Salakhutdinov and Geoffrey Hinton.
\newblock Deep boltzmann machines.
\newblock In {\em Artificial intelligence and statistics}, pages 448--455,
  2009.

\bibitem{srivastava2017veegan}
Akash Srivastava, Lazar Valkov, Chris Russell, Michael~U Gutmann, and Charles
  Sutton.
\newblock Veegan: Reducing mode collapse in gans using implicit variational
  learning.
\newblock In {\em Advances in Neural Information Processing Systems}, pages
  3308--3318, 2017.

\bibitem{starner1997real}
Thad Starner and Alex Pentland.
\newblock Real-time american sign language recognition from video using hidden
  markov models.
\newblock In {\em Motion-based recognition}, pages 227--243. Springer, 1997.

\bibitem{szegedy2016rethinking}
Christian Szegedy, Vincent Vanhoucke, Sergey Ioffe, Jon Shlens, and Zbigniew
  Wojna.
\newblock Rethinking the inception architecture for computer vision.
\newblock In {\em Proceedings of the IEEE conference on computer vision and
  pattern recognition}, pages 2818--2826, 2016.

\bibitem{terjek2019virtual}
D{\'a}vid Terj{\'e}k.
\newblock Virtual adversarial lipschitz regularization.
\newblock {\em arXiv preprint arXiv:1907.05681}, 2019.

\bibitem{turk1991face}
Matthew Turk and Alex Pentland.
\newblock Face recognition using eigenfaces.
\newblock In {\em Proceedings. 1991 IEEE computer society conference on
  computer vision and pattern recognition}, pages 586--587, 1991.

\bibitem{wang2018high}
Ting-Chun Wang, Ming-Yu Liu, Jun-Yan Zhu, Andrew Tao, Jan Kautz, and Bryan
  Catanzaro.
\newblock High-resolution image synthesis and semantic manipulation with
  conditional gans.
\newblock In {\em Proceedings of the IEEE conference on computer vision and
  pattern recognition}, pages 8798--8807, 2018.

\bibitem{warde2016improving}
David Warde-Farley and Yoshua Bengio.
\newblock Improving generative adversarial networks with denoising feature
  matching.
\newblock 2016.

\bibitem{xu1996convergence}
Lei Xu and Michael~I Jordan.
\newblock On convergence properties of the em algorithm for gaussian mixtures.
\newblock {\em Neural computation}, 8(1):129--151, 1996.

\bibitem{yang2017lr}
Jianwei Yang, Anitha Kannan, Dhruv Batra, and Devi Parikh.
\newblock Lr-gan: Layered recursive generative adversarial networks for image
  generation.
\newblock {\em arXiv preprint arXiv:1703.01560}, 2017.

\bibitem{yu2015lsun}
Fisher Yu, Ari Seff, Yinda Zhang, Shuran Song, Thomas Funkhouser, and Jianxiong
  Xiao.
\newblock Lsun: Construction of a large-scale image dataset using deep learning
  with humans in the loop.
\newblock {\em arXiv preprint arXiv:1506.03365}, 2015.

\bibitem{zhang2018self}
Han Zhang, Ian Goodfellow, Dimitris Metaxas, and Augustus Odena.
\newblock Self-attention generative adversarial networks.
\newblock {\em arXiv preprint arXiv:1805.08318}, 2018.

\bibitem{zhou2019lipschitz}
Zhiming Zhou, Jiadong Liang, Yuxuan Song, Lantao Yu, Hongwei Wang, Weinan
  Zhang, Yong Yu, and Zhihua Zhang.
\newblock Lipschitz generative adversarial nets.
\newblock {\em arXiv preprint arXiv:1902.05687}, 2019.

\end{thebibliography}
}

\end{document}